\documentclass{article} 
\usepackage[final]{colm2026_conference}
\usepackage{microtype}
\usepackage{hyperref}
\usepackage{url}
\usepackage{booktabs}
\usepackage{graphicx}
\usepackage{tabularx}
\usepackage{booktabs}
\usepackage{multirow}
\usepackage{listings}
\lstdefinestyle{promptstyle}{
  basicstyle=\ttfamily\footnotesize,
  breaklines=true,
  columns=fullflexible,
  keepspaces=true,
  showstringspaces=false
}
\usepackage[most]{tcolorbox}
\usepackage{xcolor}
\usepackage{float}
\usepackage{subcaption}

\usepackage{lineno}

\definecolor{darkblue}{rgb}{0, 0, 0.5}
\hypersetup{colorlinks=true, citecolor=darkblue, linkcolor=darkblue, urlcolor=darkblue}

\title{LLMs as Strategic Actors: Behavioral Alignment, Risk Calibration, and Argumentation Framing in Geopolitical Simulations
}

\author{\small
  Veronika Solopova, Viktoria Skorik, Maksym Tereshchenko, 
  \textbf{Alina Haidun, Ostap Vykhopen} \\
  \texttt{Correspondence: ostap.vykhopen@mantisanalytics.com}
}

\begin{document}

\ifcolmsubmission
\linenumbers
\fi

\maketitle

\begin{abstract}
Large language models (LLMs) are increasingly proposed as agents in strategic decision environments, yet their behavior in structured geopolitical simulations remains under-researched. We evaluate six popular state-of-the-art LLMs alongside results from human results across four real-world crisis simulation scenarios, requiring models to select predefined actions and justify their decisions across multiple rounds. We compare models to humans in action alignment, risk calibration through chosen actions’ severity, and argumentative framing grounded in international relations theory. Results show that models approximate human decision patterns in base simulation rounds but diverge over time, displaying distinct behavioural profiles and strategy updates. LLM explanations for chosen actions across all models exhibit a strong normative-cooperative framing centered on stability, coordination, and risk mitigation, with limited adversarial reasoning.
\end{abstract}

\section{Introduction}
Geopolitical decision-making has entered a period of heightened complexity and uncertainty. Intensifying strategic competition, hybrid conflict, technological disruption, and climate-driven instability have created overlapping crises and rapidly evolving threat environments \citep{wef2026globalrisks}. In such contexts, decision-makers must navigate incomplete information, competing objectives, and high-stakes trade-offs under time pressure.

Scenario-based simulations and crisis exercises serve as central tools for strategic preparedness. Governments, corporations, and security institutions use structured simulations to model escalation dynamics, stress-test coordination, and train adaptive reasoning. However, high-quality geopolitical simulations are resource-intensive, requiring expert facilitation and repeated participation by trained professionals. This limits their scalability and continuity.

Recent advances in large language models (LLMs) raise the possibility that AI systems could augment or simulate expert reasoning in such environments. While LLMs demonstrate increasing competence in multi-step reasoning and policy analysis, systematic comparisons between model behavior and human decision-making in realistic geopolitical simulations remain scarce. In particular, it remains unclear whether models reproduce human-like risk preferences, escalation patterns, and argumentative framing across evolving crisis contexts.

\begin{figure*}[t]
\centering
\includegraphics[width=0.8\textwidth]{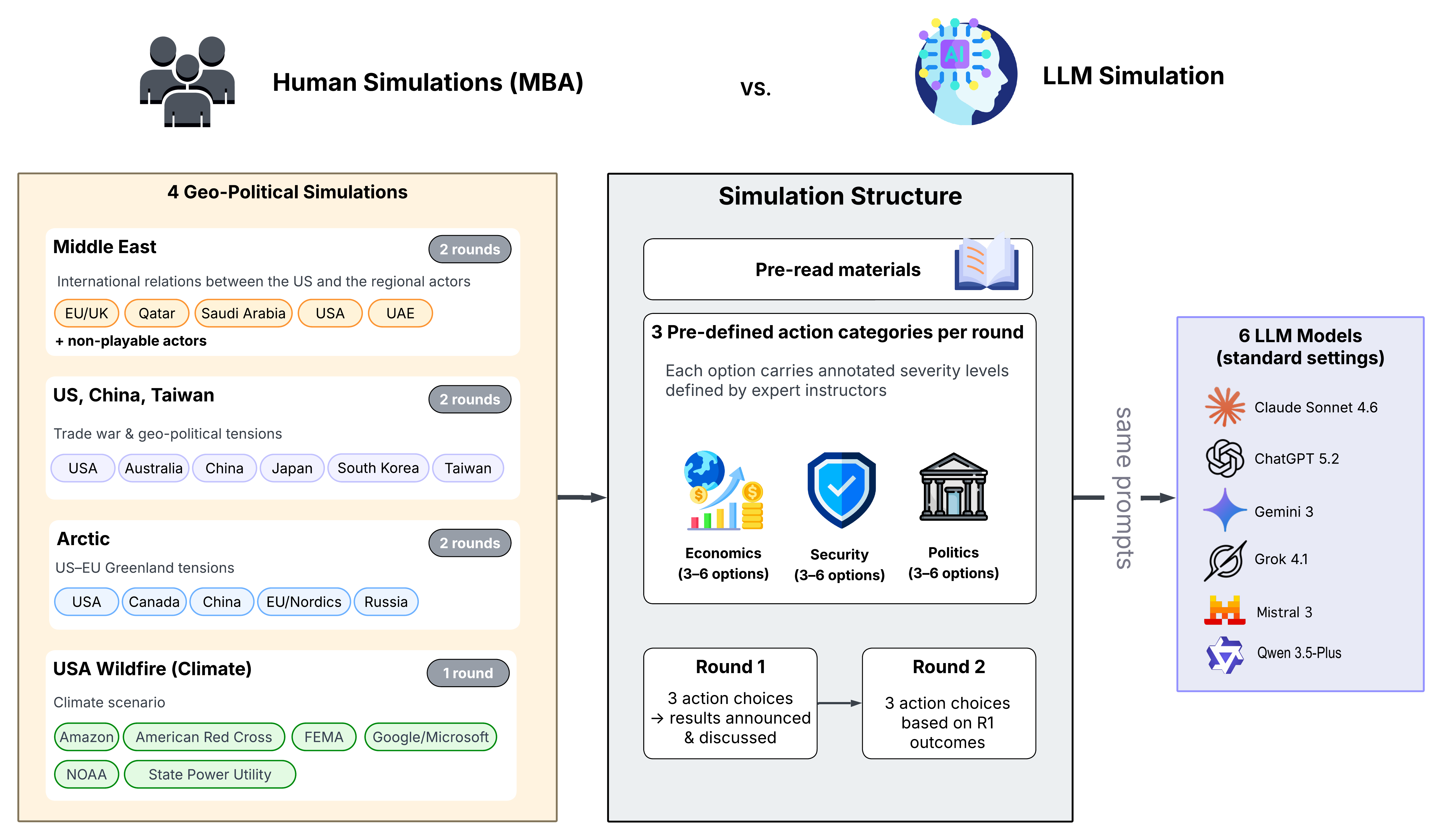}
\caption{Study design: four geopolitical simulations run with human MBA participants and six LLMs under identical prompts and structured action menus.}
\label{fig:study-design}
\end{figure*}

We address this gap by evaluating six widely used LLMs within four real-world geopolitical simulations conducted in partnership with the New Lines Institute across graduate-level policy and business programs. The scenarios span Arctic security, US–China–Taiwan tensions, Middle East dynamics, and corporate and governmental response to a U.S. wildfire crisis. Each simulation consists of structured, multi-round decision sequences in which predefined actors select among economic, security, or political actions. We replicate these simulations by providing models with the same briefings and action options as human participants. For each round, models select an action and generate a justification. We then compare model and human behavior across alignment rates, severity calibration, inter-model consistency, and argumentative framing. By grounding evaluation in structured training exercises rather than synthetic benchmarks, this study provides empirical insight into the behavioral and discursive patterns of LLMs acting as geopolitical decision-makers.

This paper makes four contributions. First, we provide one of the first empirical evaluations of multiple large language models acting as decision-makers within structured, multi-round geopolitical simulations originally designed for human training contexts. Second, we introduce a systematic framework for measuring human–AI alignment in strategic decision environments, combining exact action agreement, severity calibration, and inter-model consistency analyses. Third, we conduct a comparative examination of cross-model strategic reasoning, demonstrating how differences in framing, risk orientation, and justificatory style yield distinct behavioral profiles despite shared alignment objectives. Fourth, we propose a simulation-based methodological framework for evaluating AI systems in complex policy environments, extending assessment beyond traditional NLP benchmarks toward structured, scenario-driven decision tasks.

\section{Related Work}
Strategic simulations, wargaming exercises, and scenario-based decision training have long been used in security studies, policy education, and crisis preparedness. Such simulations allow participants to explore complex geopolitical or organizational scenarios, test responses to emerging threats, and develop decision-making skills under uncertainty \citep{perla1990wargaming, sabin2012simulatingwar}. AI tools have been introduced into these environments primarily as analytical or decision-support systems, assisting with data analysis, scenario modeling, or recommendation generation \citep{kott2015cyber}.
\subsection{LLMs as Decision-Making Agents and Political Reasoning}

Recent research has increasingly explored LLMs as planning and decision-making systems, particularly in agentic settings and multi-agent environments. Prior work has demonstrated that LLM-based agents can perform sequential reasoning, planning, and strategic coordination in synthetic tasks and simulated environments, highlighting their potential as autonomous decision-makers \citep{park2023generativeagents,zhang2025multiagent}. Benchmarks focusing on reasoning and strategy have further examined whether LLMs can approximate economic or cooperative decision-making, though such evaluations typically rely on abstract tasks rather than domain-grounded geopolitical contexts \citep{filippas2024homosilicus}.
Parallel work in political science and computational social science has examined LLM outputs in relation to ideological positioning, political framing, and policy prediction, revealing measurable tendencies toward left-wing associated preferences \citep{argyle2023outofone,motoki2024politicalbias}, which can also influence the models perception of credibility and truthfulness \citep{jakob2025polbix}. Recent studies have analyzed how LLMs frame international tensions, revealing systematic geopolitical biases and dual-framing effects across languages and model architectures \citep{guey2025mappinggeopoliticalbias11}. Similarly, discursive analyses of LLM outputs in humanitarian and geopolitical settings show that models adopt distinct rhetorical strategies depending on contextual framing and normative narratives \citep{giacalone2026discursive}.
\subsection{Generative Agents and Social Simulations}

Studies on generative agent societies have demonstrated that LLMs can simulate believable social behavior, maintain persistent identities, and interact within virtual communities \citep{park2023generativeagents}, although their simulation of humans for instance in surveys is limited \citep{cao2025surveyresponses}. Related research has explored LLM-based economic agents or behavioral replication tasks, showing that models can approximate certain patterns of human decision-making or coordination \citep{aher2023simulating,horton2023economicagents}.

More recent work has begun to examine agentic AI within geopolitical simulation environments. 
\citet{payne2026aiarmsinfluencefrontier} reports a controlled multi-turn nuclear crisis simulation in which frontier models play opposing leaders and exhibit sophisticated strategic behaviors, including deception (signal–action divergence), theory-of-mind reasoning about adversary beliefs, and metacognitive self-assessment of their own strategic capabilities. \citet{dessureault2025geopoliticalsandbox} introduce a geopolitical simulation sandbox designed as a testbed for evaluating LLM-based agents through negotiation, collaboration, and contextual decision-making within a fictitious world setting. While this work demonstrates the feasibility of using simulations to assess agentic behavior, it primarily relies on synthetic environments and predefined hierarchical agent structures rather than real-world training simulations involving human participants or constrained institutional roles.
While these approaches provide important insights into emergent behavior and social dynamics, they typically rely on open-ended environments with unconstrained actions. In contrast, real strategic simulations, particularly in geopolitical or crisis-response contexts, impose predefined roles, limited action spaces, and structured decision cycles. The present study addresses this gap by evaluating multiple LLMs within real-world geopolitical simulation scenarios conducted in educational and professional contexts.


\section{Methods}
\subsection{Data}

\paragraph{Human Data.} We collected data and settings for four geo-political simulations created and performed as part of the MBA program in political science in different universities in the US. The simulations were conducted in groups of students in 2 rounds with pre-defined possible actions, created by the expert instructors in political science. Each group of students represented one of the geo-political actors (country, union of countries or an organization) in the focus of the simulation, with 1 choice to make in each of 3 categories of actions: economic, security and political. Each type of action has from 3 to 6 options with varying annotated severity levels. After announcing and discussing the results of the first round the students had another 3 choices to make. Before the simulations the students received a short presentation to put them in the geopolitical context of the countries they will play for and readout information on each country. The actors and topics of the four scenarios are detailed in Table \ref{tab:simulations}.
\begin{table*}[t]
\centering
\caption{Overview of geopolitical simulation scenarios and participating actors.}
\label{tab:simulations}
\footnotesize
\begin{tabularx}{\textwidth}{l X X}
\toprule
\textbf{Simulation Name} & \textbf{Description} & \textbf{Actors} \\
\midrule

Middle East &
Taking place in autumn 2025, the training focused on possible scenarios of international relations development between the US and regional actors. The simulation also included non-playable regional actors with pre-defined actions (e.g., Israel, Iran, Yemen, China). &
EU/UK; Qatar; Saudi Arabia; USA; United Arab Emirates \\

US, China, Taiwan &
Taking place in late 2025, the simulation focused on next steps in the US–China trade war and escalating geopolitical tensions. &
Australia; China; Japan; South Korea; Taiwan; USA \\

Arctic &
The simulation took place in early 2026 and focused on potential next steps in intensifying tensions between the US and the EU over Greenland, as well as broader Arctic regional dynamics. &
Canada; China; EU/Nordics; Russia; USA \\

USA Wildfire (Climate) &
The only single-round simulation, focusing on corporate, NGO, and governmental responses to a simulated wildfire disaster in the United States. &
Amazon; American Red Cross; FEMA; Google and Microsoft; NOAA; State Power Utility \\

\bottomrule
\end{tabularx}
\end{table*}

\paragraph{LLM simulations}. Using the same pre-read material, and action options that were used with human participants, we generated action choices and rationale behind them for each simulation using 6 state-of-the-art LLMs chat versions: Claude Sonnet 4.6 \citep{anthropic2024claude}, ChatGPT 5.2 \citep{openai2026gpt52}, Gemini 3 \citep{gemini3flash}, Grok 4.1 \citep{grok4}, Mistral 3 \citep{mistral2026mistral3} and Qwen 3.5-Plus \citep{yang2025qwen3technicalreport} in standard settings. Prompt is available in Appendix \ref{prompt}.

\subsection{Experiments and Evaluation}

To evaluate how closely large language models reproduce human strategic behavior, we compared model-selected actions with human decisions across all simulations. Alignment was measured as exact action overlap, supplemented by micro- and macro-averaged F1 scores and pairwise agreement analyses. In addition, we examined how models calibrated risk by analyzing the severity of chosen actions across rounds and decision domains. Severity was defined on a six-level ordinal scale (Low to Extreme), enabling systematic comparison of escalation patterns between models and human participants.

To characterize explanation style, we analyzed surface-level textual properties of all explanations associated with selected actions. We measured explanation length in tokens and lexical diversity using a rolling type–token ratio (TTR) with a fixed window of 50 tokens. For shorter texts, TTR was computed over the full explanation; for longer texts, it was averaged across non-overlapping windows to stabilize estimates across varying lengths. To assess within-model phrasing variability, we further computed lexical heterogeneity as one minus the mean TF–IDF cosine similarity between all explanation pairs within each (model, simulation) subset. Explanations were vectorized using English stopword removal and 1–2-gram features. Higher values indicate greater lexical diversity, whereas lower values suggest repetitive phrasing patterns. To identify model-specific lexical signatures, we extracted TF–IDF terms that appeared in a model’s top-50 list but not in any other model’s top-50 list and visualized these unique terms.

To analyze the argumentative structure of explanations, we developed a theory-grounded framing taxonomy informed by international relations (IR) scholarship. Rather than relying on unsupervised clustering, frame categories were designed to operationalize major IR traditions—realism, liberal institutionalism, and constructivism—allowing theoretically interpretable comparison across models. Realist reasoning was captured through frames emphasizing strategic stability, deterrence, uncertainty management, and reputation signaling. Institutional reasoning encompassed multilateral cooperation and rule-based legitimacy, while constructivist reasoning included democratic values, humanitarian protection, and climate justice. This mapping enabled interpretation of explanations not only linguistically but also along established ideological dimensions.

All explanations were automatically annotated using GPT-4o under a structured prompting protocol. The model assigned one or two primary frames and up to four secondary frames from the predefined taxonomy, based solely on argumentative logic expressed in the explanation text. Prompts included explicit frame definitions and labeling rules (Appendix \ref{frame_explanations}), and outputs were constrained to a structured JSON schema to ensure reproducibility. A subset of explanations was manually reviewed by human annotators familiar with the simulation context to qualitatively verify that assigned frames aligned with the intended theoretical constructs.

Finally, to examine ideological tendencies in model reasoning, individual frames were aggregated into higher-level categories representing normative, strategic, and technical orientations. Normative framing included humanitarian, democratic, cooperative, and rule-based legitimacy frames; strategic framing encompassed deterrence, risk management, signaling, domestic legitimacy, and economic interest; technical framing captured operational feasibility. Using these groupings, we constructed a two-dimensional ideological space. The horizontal axis represents a continuum from normative/cooperative to strategic/deterrence-oriented reasoning, while the vertical axis reflects a spectrum from value-based discourse to technocratic justification. Model coordinates were computed by aggregating primary and secondary frame frequencies and projecting them into normalized positions, enabling direct comparison of framing orientation across systems.
\begin{table}[th]
\centering
\caption{Ideological grouping of framing categories and their conceptual grounding in IR theory.}
\label{tab:frame_map}
\footnotesize
\begin{tabularx}{\linewidth}{l l X}
\toprule
\textbf{Orientation} & \textbf{Frame} & \textbf{Conceptual Meaning} \\
\midrule
\multirow{5}{*}{Normative / Cooperative}
& Humanitarian Protection & Civilian safety, harm reduction \\
& Democratic Values \& Rights & Democracy, liberal norms \\
& Multilateralism, Cooperation & Alliances, institutions \\
& Climate, Justice, Equity & Equity, global responsibility \\
& Legal Normative Order & International law, legitimacy \\
\addlinespace
\multirow{5}{*}{Strategic / Deterrence}
& Strategic\_Stability\_Deescalation & Deterrence, balance of power \\
& Reputation Signaling & Credibility, prestige \\
& Risk Uncertainty & Threat assessment, mitigation \\
& Domestic Legitimacy & Internal political constraints \\
& Economic Resource & Material costs, markets \\
\addlinespace
Technical / Neutral
& Technical, Operational & Implementation, logistics \\
\bottomrule
\end{tabularx}
\end{table}
\section{Results}

\subsection{Models’ and Human Actions}
\begin{table}[t]
\centering
\caption{Alignment of model-selected actions with human participant decisions across simulation rounds (Round 1: $n=63$, Round 2: $n=45$). Alignment is reported as micro-F1 (equivalent to exact decision agreement under single-label action selection) and macro-F1 (averaged across action categories).}
\label{tab:alignment}
\footnotesize
\begin{tabular}{lcccc}
\toprule
 & \multicolumn{2}{c}{\textbf{Round 1}} & \multicolumn{2}{c}{\textbf{Round 2}} \\
\cmidrule(lr){2-3} \cmidrule(lr){4-5}
\textbf{Model} & \textbf{F1-micro} & \textbf{F1-macro} & \textbf{F1-micro} & \textbf{F1-macro} \\
\midrule
Gemini  & 0.540 & 0.405 & 0.333 & 0.276 \\
Claude  & 0.533 & 0.480 & 0.262 & 0.153 \\
Mistral & 0.476 & 0.364 & 0.156 & 0.125 \\
ChatGPT & 0.460 & 0.328 & 0.333 & 0.265 \\
Qwen    & 0.317 & 0.247 & 0.267 & 0.202 \\
Grok    & 0.254 & 0.162 & 0.225 & 0.138 \\
\bottomrule
\end{tabular}
\end{table}

\begin{figure}[t]
    \centering
    \begin{subfigure}[t]{0.48\linewidth}
        \centering
        \includegraphics[width=\linewidth]{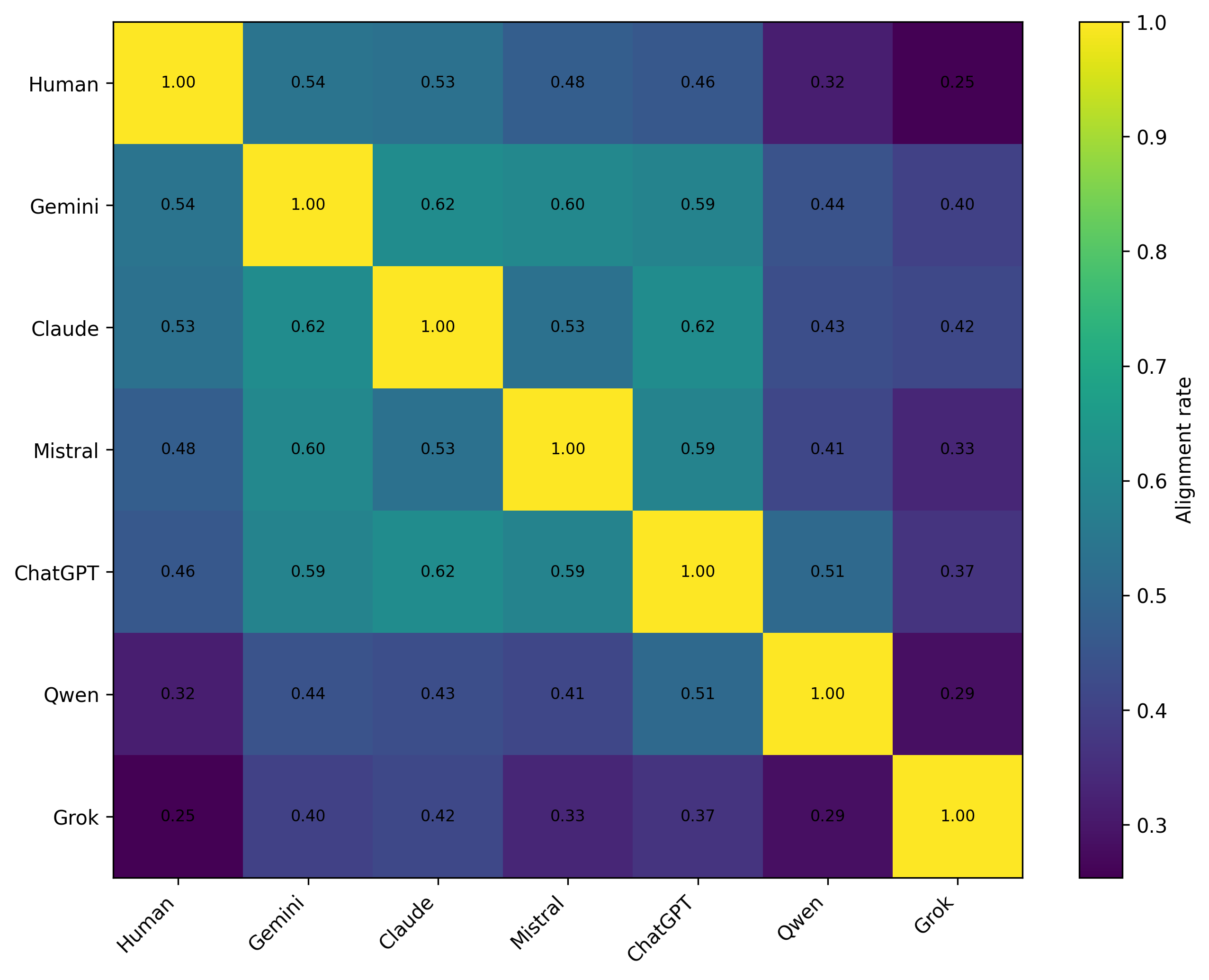}
        \caption{Round 1}
        \label{fig:align-r1}
    \end{subfigure}\hfill
    \begin{subfigure}[t]{0.48\linewidth}
        \centering
        \includegraphics[width=\linewidth]{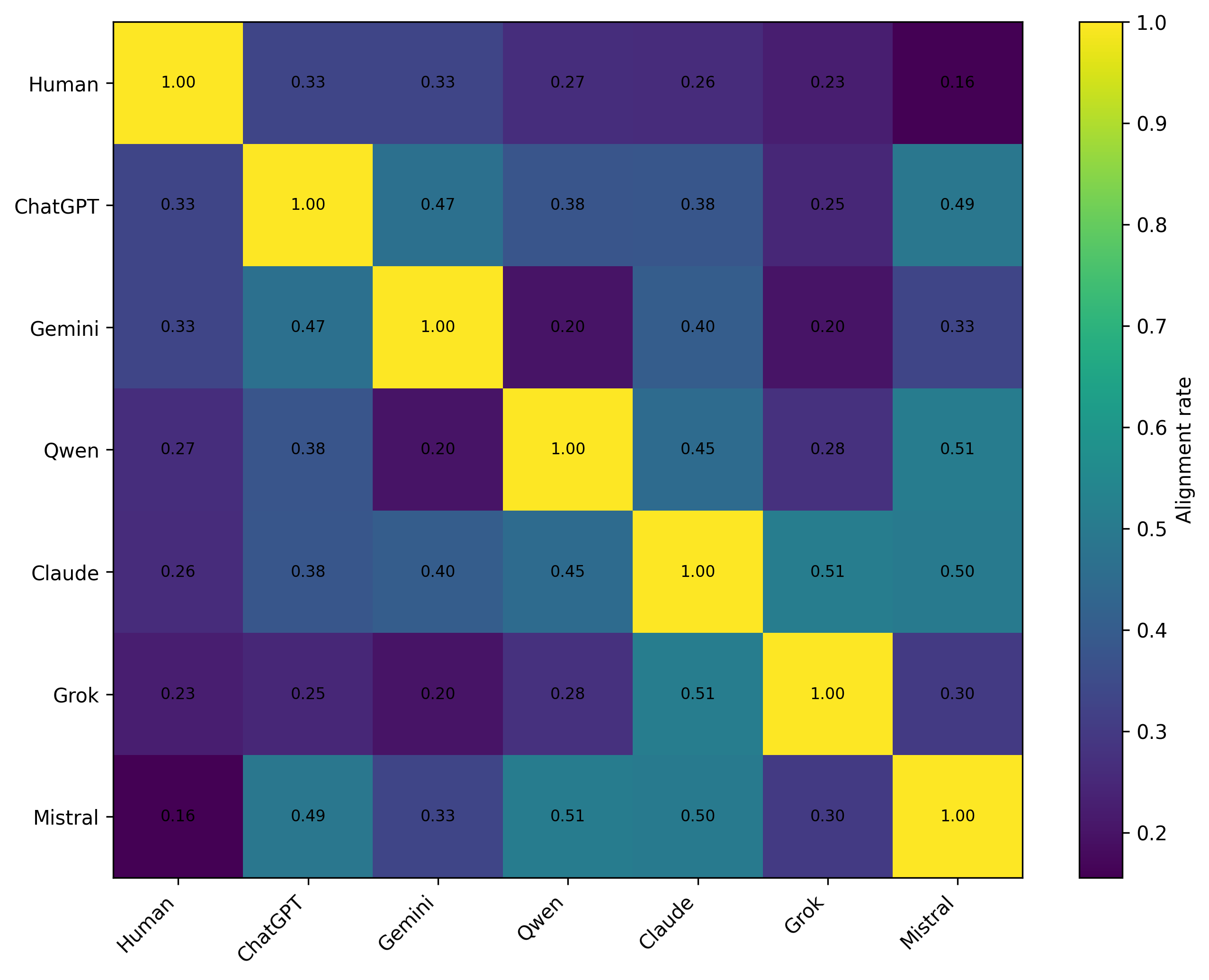}
        \caption{Round 2}
        \label{fig:align-r2}
    \end{subfigure}
    \caption{Pairwise decision alignment among humans and LLMs by round. Each cell reports the fraction of shared questions for which two agents selected the same action (exact match). Models are ordered by alignment with humans within each round.}
    \label{fig:pairwise-alignment}
\end{figure}
Models show moderate alignment with human participants. In Round 1, agreement ranges from approximately 0.25 to 0.54, with Gemini and Claude demonstrating the highest similarity to human decisions, followed by Mistral and ChatGPT, while Qwen and Grok exhibit lower alignment. This suggests that several models capture baseline strategic heuristics observed in human decision-making, though considerable variability remains.

Inter-model comparison reveals a relatively coherent behavioral cluster. Krippendorff’s $\alpha$ reaches 0.41 among models, indicating moderate internal consistency, but declines slightly to 0.39 when human decisions are included. Thus, LLMs align marginally more strongly with one another than with human participants, pointing to shared training priors or reasoning templates that partially diverge from expert judgment. Pairwise heatmaps (Figure~\ref{fig:pairwise-alignment}) reinforce this structure: Claude, Gemini, ChatGPT, and Mistral form a central alignment cluster, whereas Grok and Qwen consistently appear more peripheral. Importantly, alignment reflects similarity in action selection rather than normative quality or strategic optimality. While several models exceed 0.5 agreement with humans in Round 1, Round 2 alignment drops to approximately 0.16–0.33. Inter-model agreement similarly fragments in later rounds, indicating increasing heterogeneity as scenarios evolve and contextual complexity accumulates. This temporal divergence suggests that initial convergence may stem from shared default heuristics, whereas extended strategic reasoning amplifies differences in model updating and interpretation.

Alignment also varies by simulation context. The wildfire (climate) scenario produces the highest human–model agreement, whereas the Arctic simulation yields the lowest, with the Middle East and US–China–Taiwan cases occupying an intermediate and closely clustered range. This pattern indicates that model–human convergence is partly scenario-dependent, with cooperative or crisis-response settings eliciting greater similarity than competitive geopolitical environments.

\subsection{Action nature and severity}
\begin{figure*}[t]
    \centering
    \begin{subfigure}[t]{0.49\textwidth}
        \centering
        \includegraphics[width=\linewidth]{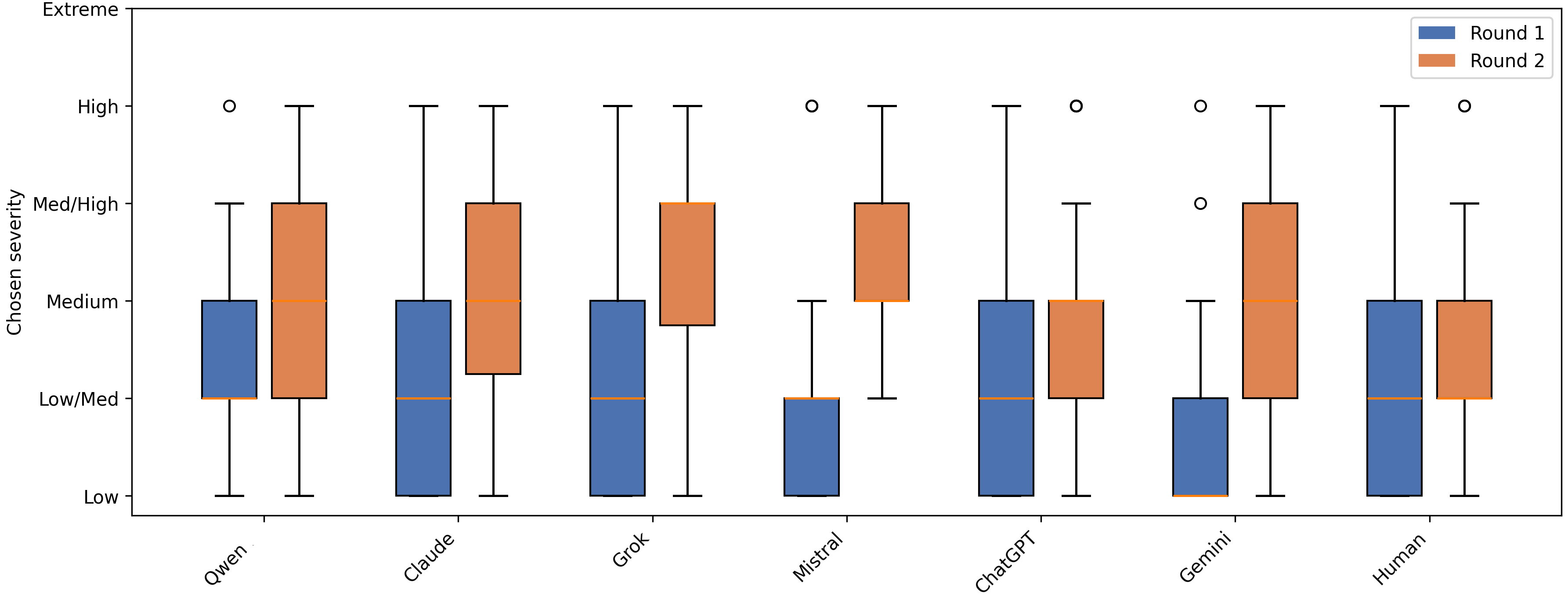}
        \caption{Severity distribution by round.}
        \label{fig:severity-round}
    \end{subfigure}
    \hfill
    \begin{subfigure}[t]{0.496\textwidth}
        \centering
        \includegraphics[width=\linewidth]{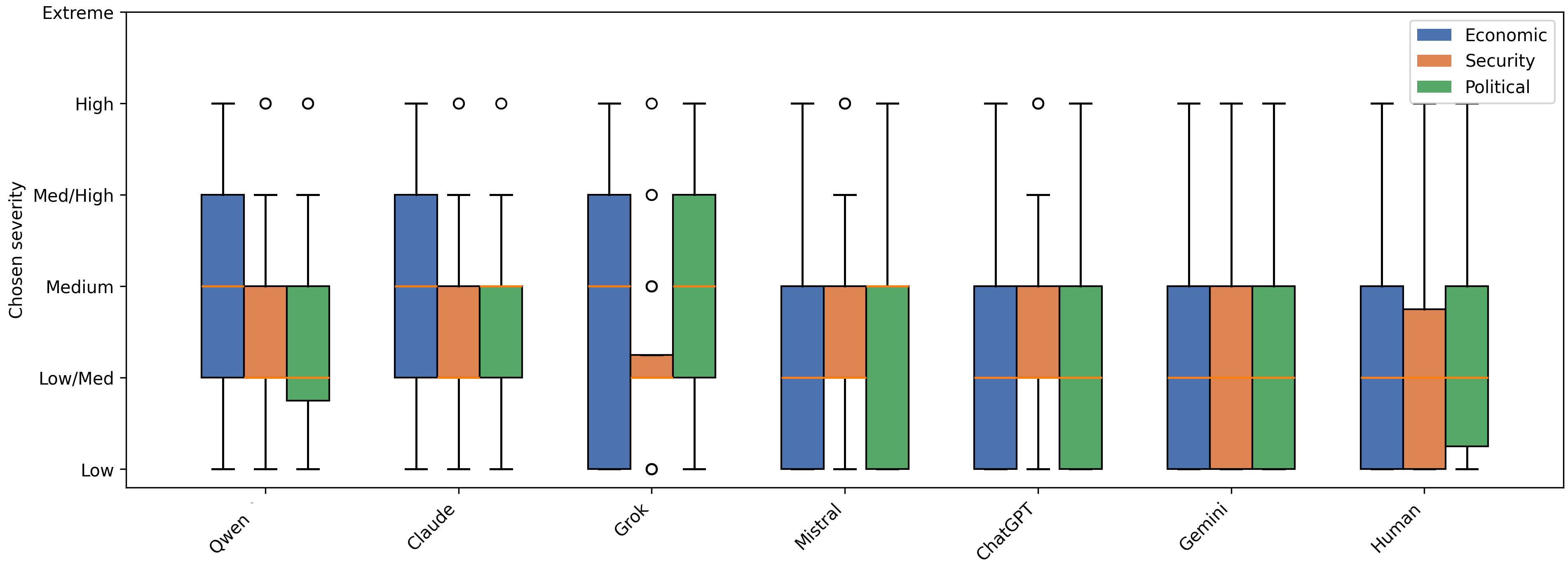}
        \caption{Severity distribution by decision type.}
        \label{fig:severity-dimension}
    \end{subfigure}
    \caption{ Distribution of chosen action severity across models and human participants. (a) severity distributions by round (b) severity distributions by decision dimension (Economic, Security, Political/Diplomatic).}
    \label{fig:severity}
\end{figure*}
Across models and human participants, severity distributions shift upward from Round 1 to Round 2, indicating systematic escalation as scenarios unfold. Median values increase in nearly all cases, suggesting adaptive responses to evolving context rather than fixed risk preferences. However, the magnitude of escalation varies. Gemini and Claude show the strongest upward shifts, Mistral displays a clear but more concentrated transition, and ChatGPT remains comparatively stable. Qwen and Grok exhibit broader variability across scenarios, while human participants occupy a moderate position, escalating without substantial changes in distribution width. Mann–Whitney U tests confirm that severity increases significantly between rounds for all models after Benjamini–Hochberg correction (all p\_{fdr} \textless .001), with large effect sizes ( r = .45–.76). Mistral demonstrates the strongest escalation effect, whereas Qwen shows the weakest. These results indicate that round progression systematically alters risk calibration rather than reflecting random variation.

Severity differences across decision domains are limited. Although economic actions display slightly higher medians and broader upper ranges, security decisions show tighter distributions, and political actions exhibit greater inter-model dispersion, Kruskal–Wallis tests reveal no statistically significant domain effects after correction. Pairwise comparisons likewise show minimal differentiation, with only Grok displaying a significant difference between Security and Political decisions (p\_{fdr} = .049, r = .40). Thus, severity adjustment appears driven more by temporal dynamics than by domain-specific reasoning.

Token-level analysis further clarifies these patterns. In political decisions, most models closely mirror human participants by favoring coordination-oriented actions (e.g., diplomatic, summit, coordinate), with Gemini emphasizing procedural diplomacy (envoy, officials) and Grok adopting a more transactional tone (trade, delegation). Economic decisions show clearer divergence: Humans, ChatGPT, and Gemini prioritize alignment-oriented measures, whereas Claude, Mistral, and Qwen emphasize sanctions and supply-chain interventions, indicating a more coercive framing. In the security domain, vocabulary converges strongly across all actors, with repeated reliance on operational and surveillance terms (conduct, military, intelligence).

\section{Models’ rationale}

\begin{figure*}[t]
    \centering
    \begin{subfigure}[t]{0.4\textwidth}
        \centering
        \includegraphics[width=\linewidth]{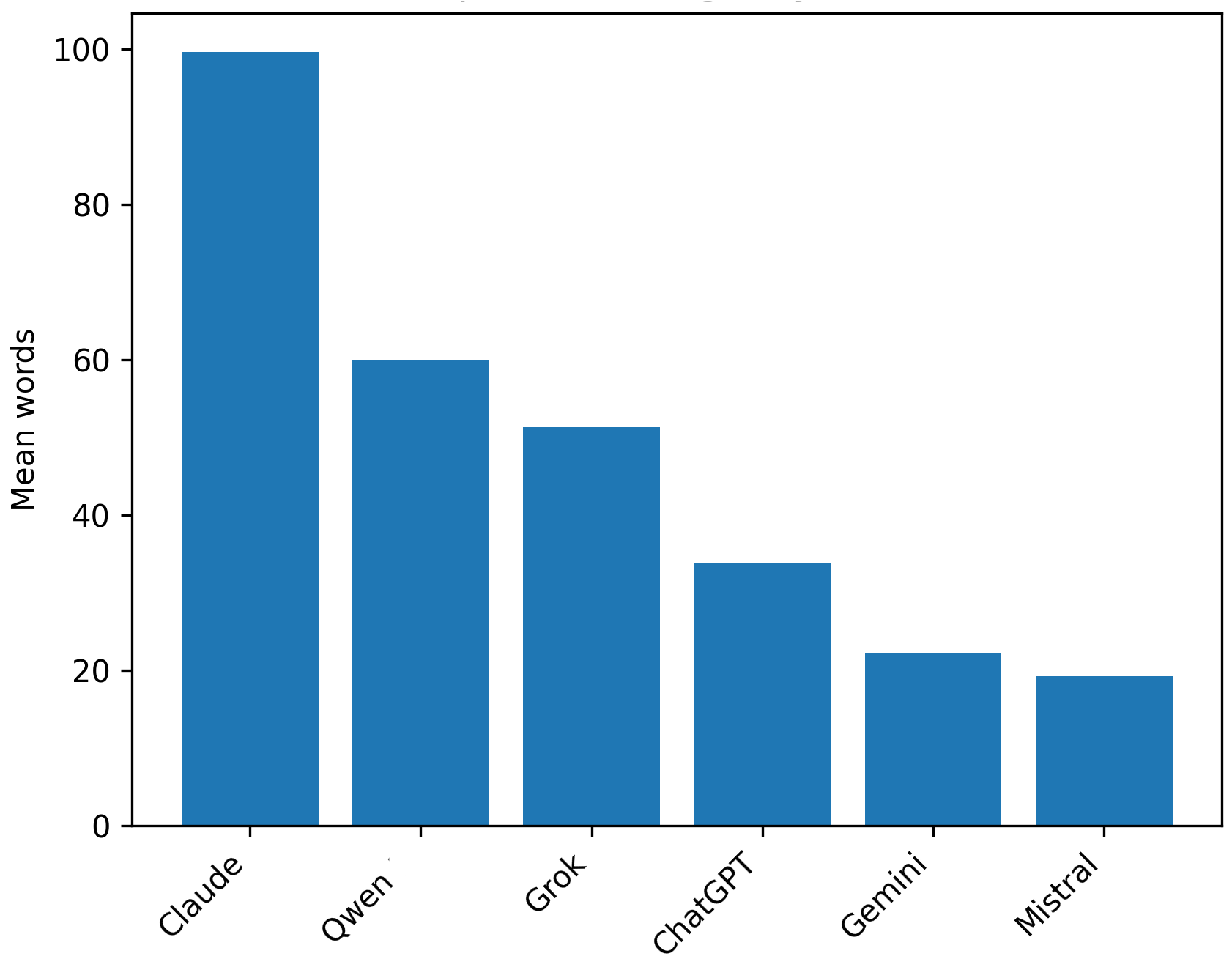}
        \caption{Mean explanation length (words) by model.}
        \label{fig:explanation-length}
    \end{subfigure}
    \hfill
    \begin{subfigure}[t]{0.55\textwidth}
        \centering
        \includegraphics[width=\linewidth]{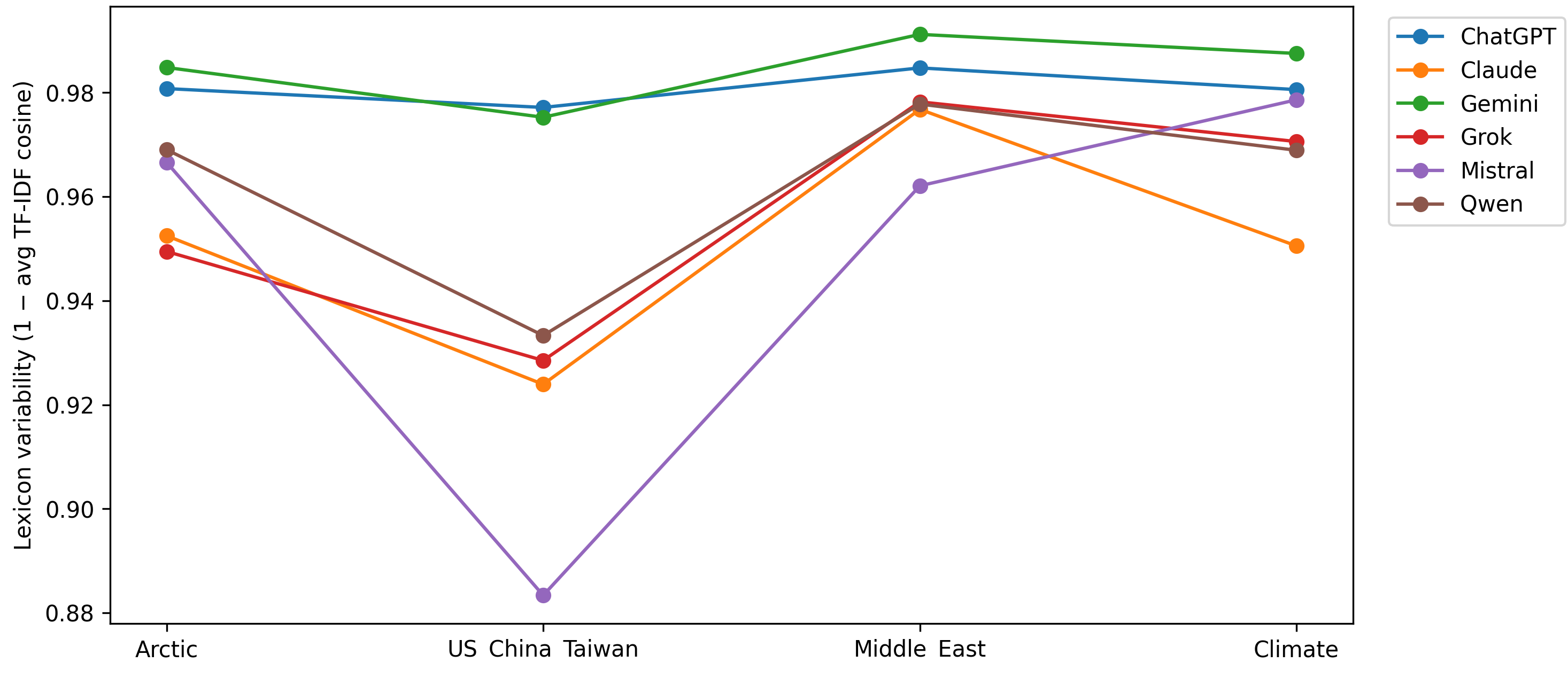}
        \caption{Within-model lexical variability across simulations (1 - mean TF-IDF cosine similarity).}
        \label{fig:lexicon-variability}
    \end{subfigure}
    \caption{Surface-level characteristics of model explanations. Explanation length reflects verbosity differences across models, while lexical variability captures within-model repetition versus linguistic diversity across simulations.}
    \label{fig:lexical-analysis}
\end{figure*}
Explanation length varies substantially across models (Figure \ref{fig:explanation-length}). Claude produces markedly longer justifications ($\approx$100 tokens on average), followed by Qwen ($\approx$60) and Grok ($\approx$50). ChatGPT occupies a mid-range ($\approx$30–35), whereas Gemini and Mistral generate comparatively concise explanations ($\approx$20–25). These differences indicate distinct verbosity profiles that are largely model-specific rather than scenario-driven.

Lexical variability within models is generally low across simulations (Figure \ref{fig:lexicon-variability}), indicating substantial within-model consistency in phrasing. However, topic-sensitive shifts emerge. Most models maintain relatively stable variability across scenarios, whereas Mistral shows the clearest contextual dependence, with a pronounced reduction in lexical diversity in the “US–China–Taiwan” simulation. This suggests that certain geopolitical contexts may induce more template-like or repetitive explanatory structures in specific systems.

Model-specific frequent non-overlapping tokens analysis (Figure \ref{fig:unique-lexicon}) further reveal systematic stylistic differentiation. ChatGPT is characterized by security–maritime vocabulary (e.g., maritime, deterrence, cyber), Claude emphasizes institutional and alliance-oriented language (e.g., commitment, NATO, framework), and Gemini foregrounds regional actors and protection-related terms (e.g., South Korea, Beijing, protection). Qwen incorporates geographically and economically anchored tokens (e.g., Greenland, semiconductor, market, wildfires), while Grok favors analytical terminology (e.g., constraints, objectives), and Mistral highlights escalation-oriented language (e.g., tensions, escalate, pressure).

\subsubsection{Explanation Framing analysis}
\begin{figure*}[t]
    \centering

    \begin{subfigure}[t]{0.60\linewidth}
        \centering
        \includegraphics[width=\linewidth]{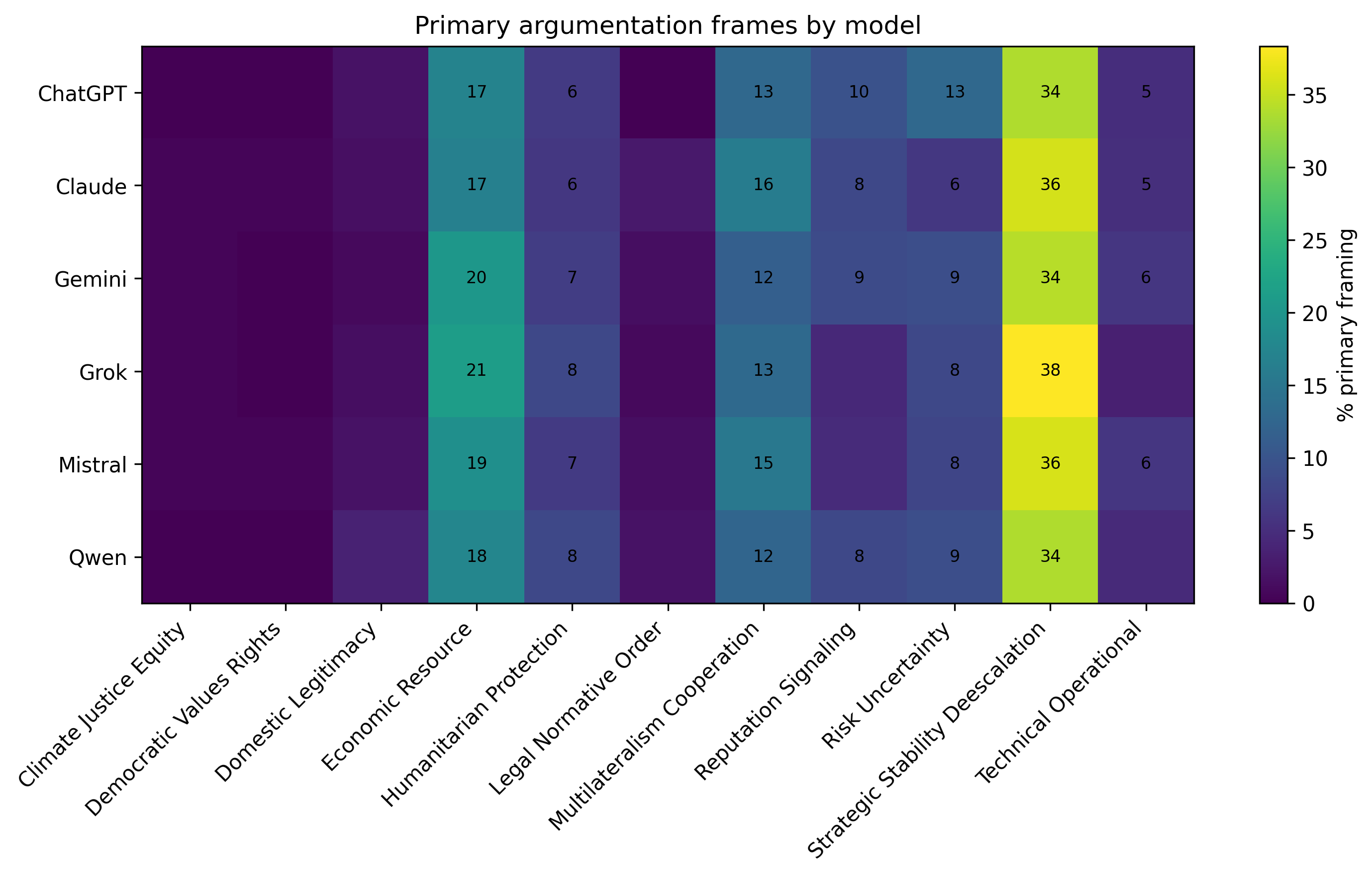}
        \caption{Primary argumentation frames by model (percentage shares).}
        \label{fig:primary-framing-heatmap}
    \end{subfigure}
    \hfill
    \begin{subfigure}[t]{0.38\linewidth}
        \centering
        \includegraphics[width=\linewidth]{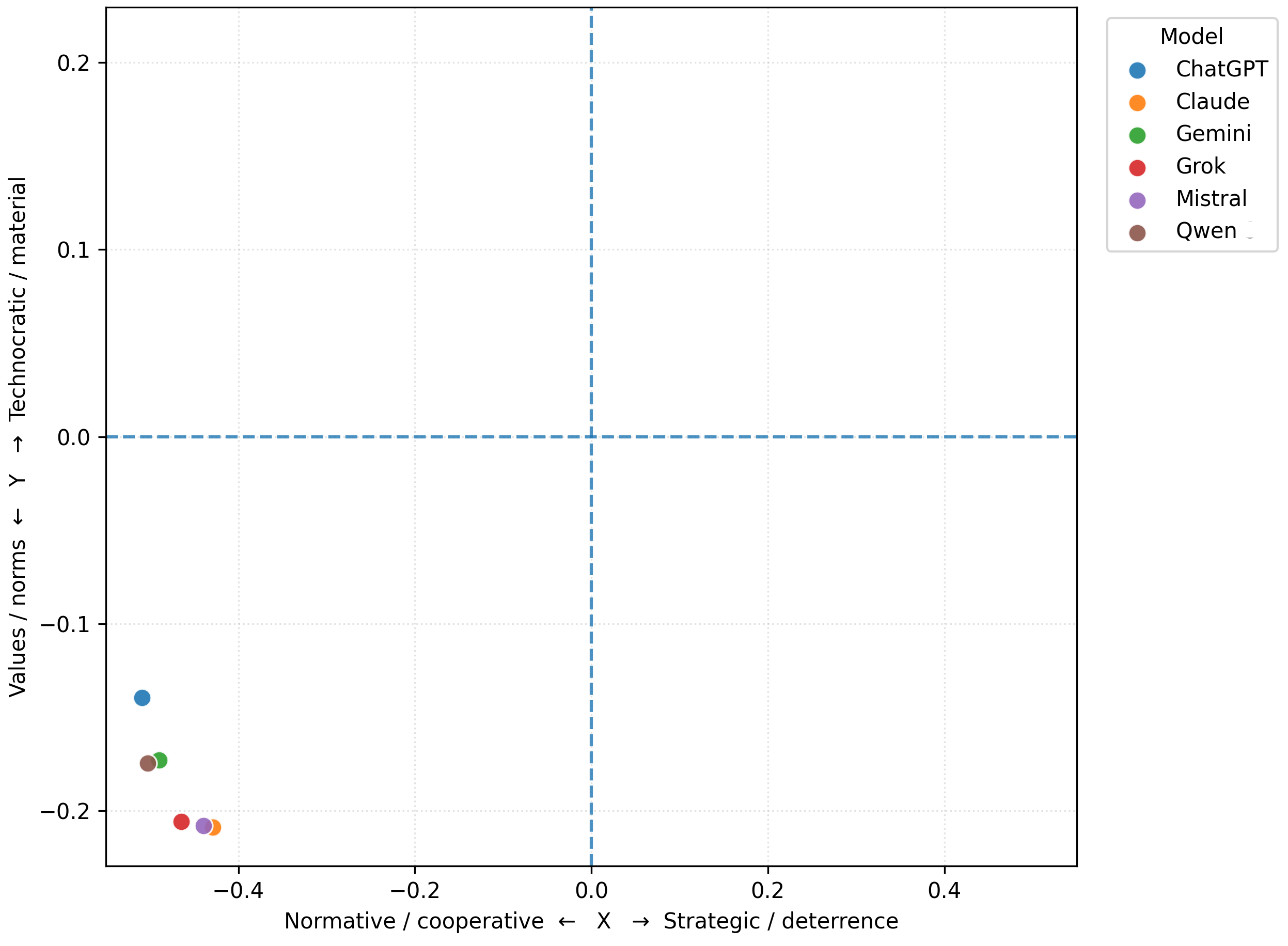}
        \caption{2D ideological projection of framing (primary + secondary).}
        \label{fig:framing-2d-map}
    \end{subfigure}

    \caption{Framing structure and ideological orientation across models. 
    (a) Exact primary frame shares per model. 
    (b) Aggregate ideological projection derived from grouped primary and secondary frames. 
    All models cluster within the normative–cooperative region, with variation primarily in degree rather than direction of strategic orientation.}
    
    \label{fig:framing-overview}
\end{figure*}
\paragraph{Primary frames.} Across all models, Strategic Stability \& De-escalation dominates primary framing, accounting for roughly one-third of explanations ($\approx$34–38\%). Grok shows the strongest reliance ($\approx$38\%), while ChatGPT, Gemini, and Qwen cluster around $\approx$34\%, indicating a shared stabilization-oriented posture. This convergence suggests that, despite architectural differences, models default to conflict-avoidance and risk-mitigation logic rather than escalation narratives.
The second most frequent primary frame is Economic \& Resource Logic ($\approx$17–21\%), with Grok again highest and ChatGPT and Claude slightly lower. Multilateralism \& Cooperation forms a third pillar ($\approx$12–16\%), particularly prominent in Claude, consistent with its alliance- and institution-focused lexical profile. Other frames appear comparatively rarely: Humanitarian Protection, Risk \& Uncertainty, and explicitly normative frames such as Democratic Values \& Rights or Climate Justice \& Equity each remain below 10\%, indicating that primary reasoning is predominantly strategic and instrumental rather than value-driven.
\paragraph{Secondary frames.} Secondary framing reveals greater stylistic differentiation. Risk \& Uncertainty is the most common secondary frame ($\approx$15–24\%), especially for ChatGPT and Qwen, reflecting frequent hedging or contingency reasoning. Claude and Mistral rely more on Reputation \& Signaling ($\approx$17–22\%), emphasizing credibility and international perception. Domestic Legitimacy appears more often as a secondary than primary frame ($\approx$13–21\%), particularly for Grok and Claude, suggesting that internal political considerations are embedded within broader strategic narratives. Legal and humanitarian frames remain marginal even at the secondary level.
\paragraph{Ideological projection.} Projecting frames onto the normative–strategic axis shows that all models occupy the normative–cooperative region (Figure \ref{fig:framing-2d-map}). Negative axis values across systems indicate consistent preference for diplomacy, coordination, and stability-oriented justification rather than deterrence or coercive positioning. This reflects the dominance of de-escalatory and multilateral frames across models.
Relative differences nevertheless emerge along a gradient of normative intensity. Qwen and ChatGPT exhibit the strongest cooperative orientation; Grok and Claude display comparatively more strategic elements through capability and signaling language; Gemini and Mistral occupy intermediate positions. However, variation occurs within a shared discourse space rather than across distinct ideological camps.

\section{Discussion and Conclusion}
This study shows that contemporary large language models can approximate human decision patterns in structured geopolitical simulations, yet they do not converge on a single “human-like” strategy. Instead, models exhibit partially overlapping but distinct behavioral profiles that vary across rounds, decision domains, and framing styles. At the level of action selection, Gemini and Claude demonstrate the strongest alignment with human participants, particularly in the first round, suggesting reliance on broadly shared strategic heuristics. ChatGPT and Mistral occupy an intermediate position, while Qwen and Grok diverge more consistently from human choices. Notably, alignment declines across nearly all models in later rounds, indicating that as contextual complexity accumulates, models update strategies differently from humans and from one another.
Severity patterns reinforce this dynamic. Most models escalate from Round 1 to Round 2 in a manner similar to human participants, but the magnitude of adjustment varies: Gemini and ChatGPT follow more gradual, human-like trajectories, whereas Mistral and Grok exhibit sharper shifts, and Qwen remains comparatively stable. While economic decisions tend to involve higher-severity actions and political decisions cluster around moderate diplomacy, statistical tests suggest that temporal progression exerts a stronger influence on severity calibration than domain-specific factors.
Framing analysis reveals even stronger convergence. Across systems, explanations consistently occupy a normative–cooperative region of the ideological space, emphasizing de-escalation, multilateral coordination, and stability-oriented reasoning. Although strategic-realist frames such as deterrence, signaling, and risk management are present, models rarely justify actions through explicit adversarial or power-balancing narratives. Differences between systems reflect variation in emphasis rather than distinct ideological positions: ChatGPT and Qwen lean more strongly toward cooperative framing, Claude and Grok incorporate somewhat more strategic signaling, and Gemini and Mistral occupy intermediate positions.

Taken together, the findings suggest that LLMs operationalize international relations logic through a broadly stabilizing and risk-averse lens. They can function as plausible simulation actors, approximating human decision patterns while exhibiting consistent discursive biases. However, their shared normative orientation may limit their ability to represent the full spectrum of geopolitical reasoning, particularly adversarial or explicitly power-centric perspectives. Future work should examine more competitive or high-uncertainty scenarios, as well as human–AI hybrid settings, to better understand how AI agents shape strategic reasoning in complex policy environments.



\bibliography{colm2026_conference}
\bibliographystyle{colm2026_conference}
\newpage{}
\appendix
\section{Appendix}

\subsubsection{Prompt used for all models}
\label{prompt}
\begin{tcolorbox}[
  colback=gray!10,
  colframe=gray!40,
  boxrule=0.5pt,
  arc=2pt,
  left=6pt,right=6pt,top=6pt,bottom=6pt,
  title=Prompt (verbatim),
  fonttitle=\bfseries,
]
\footnotesize\ttfamily
Prompt: You are an experienced geopolitical intelligence analyst and you work in the big global think tank and your focus and expertise is geopolitical simulations. Now your task is to conduct a simulation. Please, first of all, review pre-read materials for this simulation. <Attach Preread Materials>

Also, please review the list of the incidents which happened in the past for the actors which participate in simulation (<Simulation Actors>). That is needed just to review the geopolitical situation before the simulation. <Attach List Of Incidents>

Country Profiles: In this simulation we have 5 government actors. In the attached files please review the country profiles of the <Simulation Actors>. <Attach Country Profiles>

Now review attached pre-simulation incidents. Based on the all provided information, you need to make decisions for other five government actors: <Simulation Actors>. For every government actor, you have two rounds. In every round, you need to make three decisions, one per category: economic hybrid action, security intel action, and diplomatic political action.

For every decision, you need to provide detailed context for why you are taking such kinds of decisions. So basically, we should have around 15 decisions for one round and 15 decisions for the second round (three decisions per country per round).

Please review the attached file and make decisions for all five countries. You need to take decisions from the provided options in the last xlsx file, specified in the column "Title". Please choose decisions only from provided options and make it for each of the actors.

Please structure output as:
\{
  "Actor x": \{
    "round\_1": \{
      "choice\_1": \{ "title": "......", "explanation": "......" \},
      "choice\_2": \{ "title": "......", "explanation": "......" \},
      "choice\_3": \{ "title": "......", "explanation": "......" \}
    \},
    "round\_2": \{
      "choice\_1": \{ "title": "......", "explanation": "......" \},
      "choice\_2": \{ "title": "......", "explanation": "......" \},
      "choice\_3": \{ "title": "......", "explanation": "......" \}
    \}
  \}
\}
\end{tcolorbox}
\subsection{Lexical Analysis of the model's rationale}
\begin{figure}[th]
    \centering
    \includegraphics[width=\textwidth]{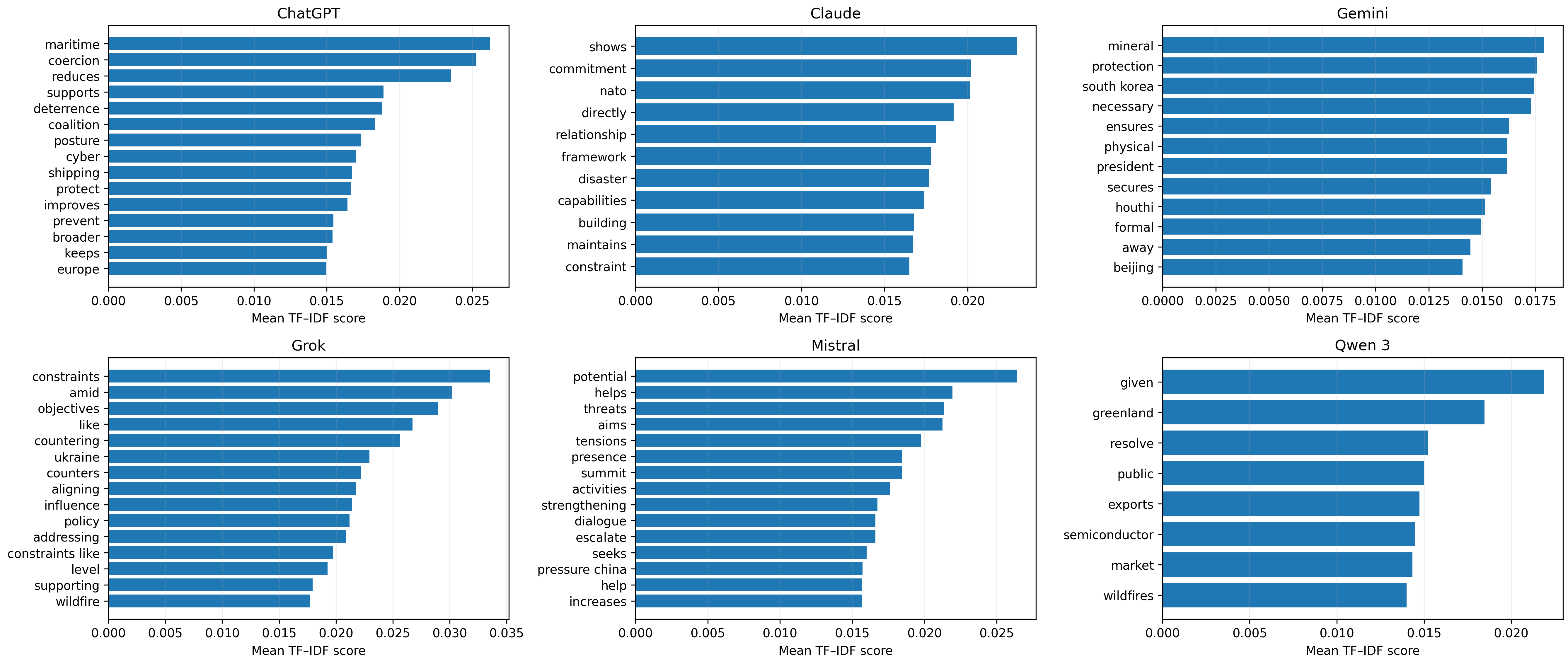}
    \caption{Model-specific lexical signatures. Terms shown represent tokens that appear within a model’s top-50 TF–IDF features but not in the top-50 of any other model. Bars indicate mean TF–IDF weight across explanations.}
    \label{fig:unique-lexicon}
\end{figure}
\begin{table*}[p]
\centering
\caption{Top 10 most frequent action tokens per model and decision dimension (stopwords removed).}
\label{tab:top-tokens}
{\scriptsize
\setlength{\tabcolsep}{4pt}
\renewcommand{\arraystretch}{1.05}

\begin{tabular}{lll}
\toprule
\textbf{Model} & \textbf{Dimension} & \textbf{Top Tokens} \\
\midrule

\multirow{3}{*}{Human}
& Economic & economic, alignment, exports, deepen, asci, china, against, allies, chinese, entities \\
& Political & diplomatic, summit, propose, coordinate, allied, chinese, iran, organize, phone, response \\
& Security & conduct, arctic, military, surveillance, activities, exercises, strait, intelligence, reconnaissance, satellite \\

\midrule

\multirow{3}{*}{ChatGPT}
& Economic & exports, economic, russian, alignment, chinese, affiliated, china, korea, against, campaign \\
& Political & diplomatic, propose, summit, response, call, coordinate, iran, organize, phone, chinese \\
& Security & conduct, military, arctic, reconnaissance, surveillance, china, vessels, deploy, intelligence, strait \\

\midrule

\multirow{3}{*}{Claude}
& Economic & exports, against, russian, asci, china, chinese, sanctions, companies, economic, facilitate \\
& Political & diplomatic, propose, summit, call, chinese, coordinate, delegation, emergency, allied, dispatch \\
& Security & conduct, military, exercises, china, arctic, operations, sea, south, allied, east \\

\midrule

\multirow{3}{*}{Gemini}
& Economic & exports, chinese, economic, alignment, china, deepen, russian, supply, affiliated, against \\
& Political & diplomatic, coordinate, envoy, call, chinese, organize, phone, special, allied, officials \\
& Security & activities, surveillance, arctic, conduct, military, intelligence, satellite, deploy, drone, exercises \\

\midrule

\multirow{3}{*}{Mistral}
& Economic & exports, russian, chinese, sanctions, economic, oil, supply, china, impose, against \\
& Political & diplomatic, propose, summit, coordinate, call, chinese, organize, phone, emergency, iran \\
& Security & conduct, reconnaissance, surveillance, arctic, china, deploy, military, against, intelligence, naval \\

\midrule

\multirow{3}{*}{Qwen}
& Economic & exports, russian, against, china, supply, chinese, companies, economic, sanctions, alignment \\
& Political & diplomatic, call, chinese, iran, response, summit, coordinate, officials, organize, phone \\
& Security & conduct, military, arctic, china, exercises, reconnaissance, surveillance, naval, intelligence, satellite \\

\midrule

\multirow{3}{*}{Grok}
& Economic & chinese, exports, russian, supply, china, against, asci, campaign, companies, coordinate \\
& Political & diplomatic, trade, china, delegation, coordinate, envoy, iran, response, special, allied \\
& Security & conduct, military, exercises, arctic, china, surveillance, sea, south, intelligence, east \\

\bottomrule
\end{tabular}
}
\end{table*}
\clearpage
\subsection{Framing and alignment analysis}

\begin{figure*}[h]
\centering

\begin{subfigure}[t]{0.48\linewidth}
    \centering
    \includegraphics[width=0.85\linewidth]{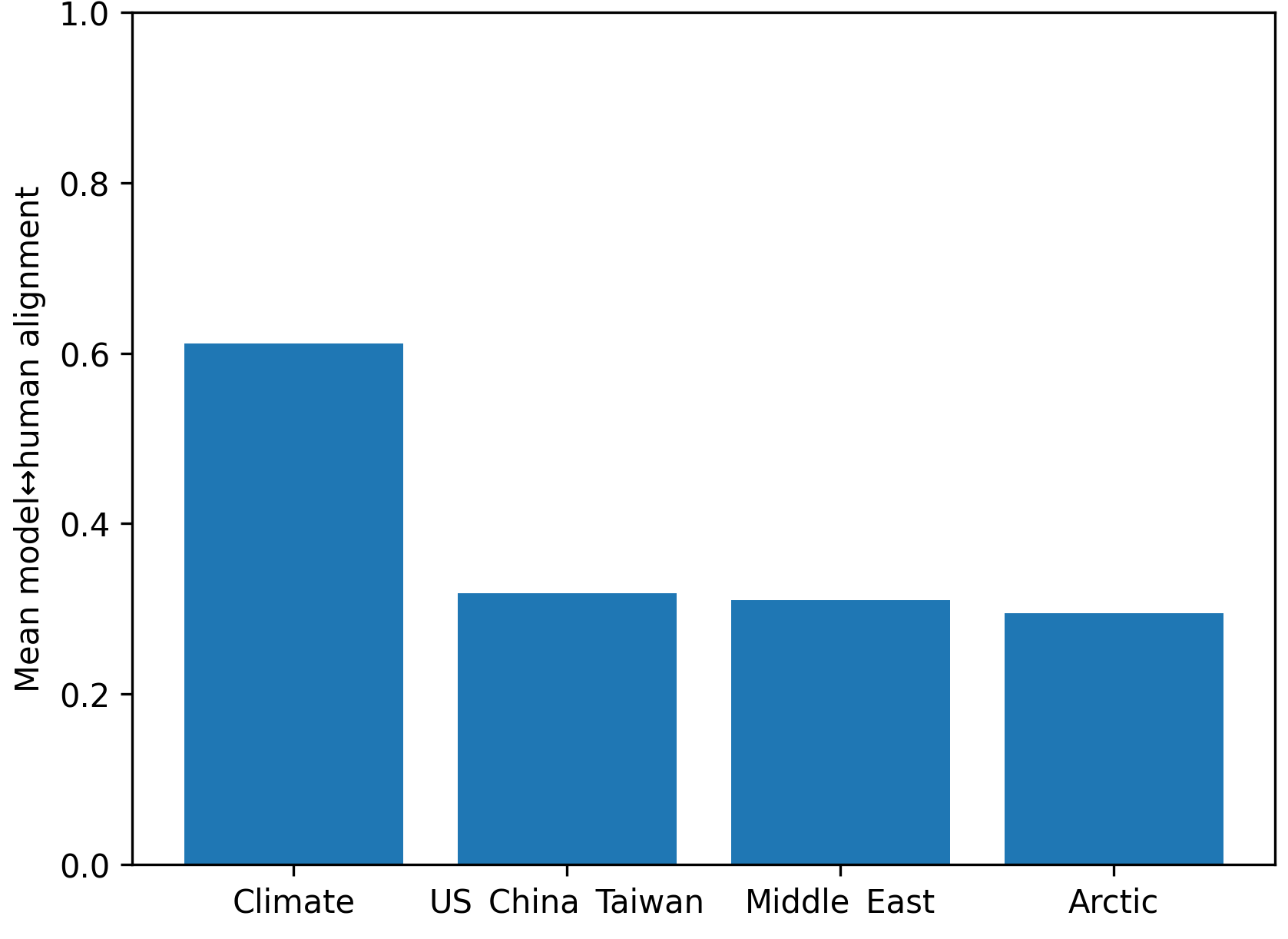}
    \caption{Mean human-model alignment by simulation (micro-F1 / exact agreement).}
    \label{fig:alignment-by-simulation}
\end{subfigure}
\hfill
\begin{subfigure}[t]{0.48\linewidth}
    \centering
    \includegraphics[width=0.85\linewidth]{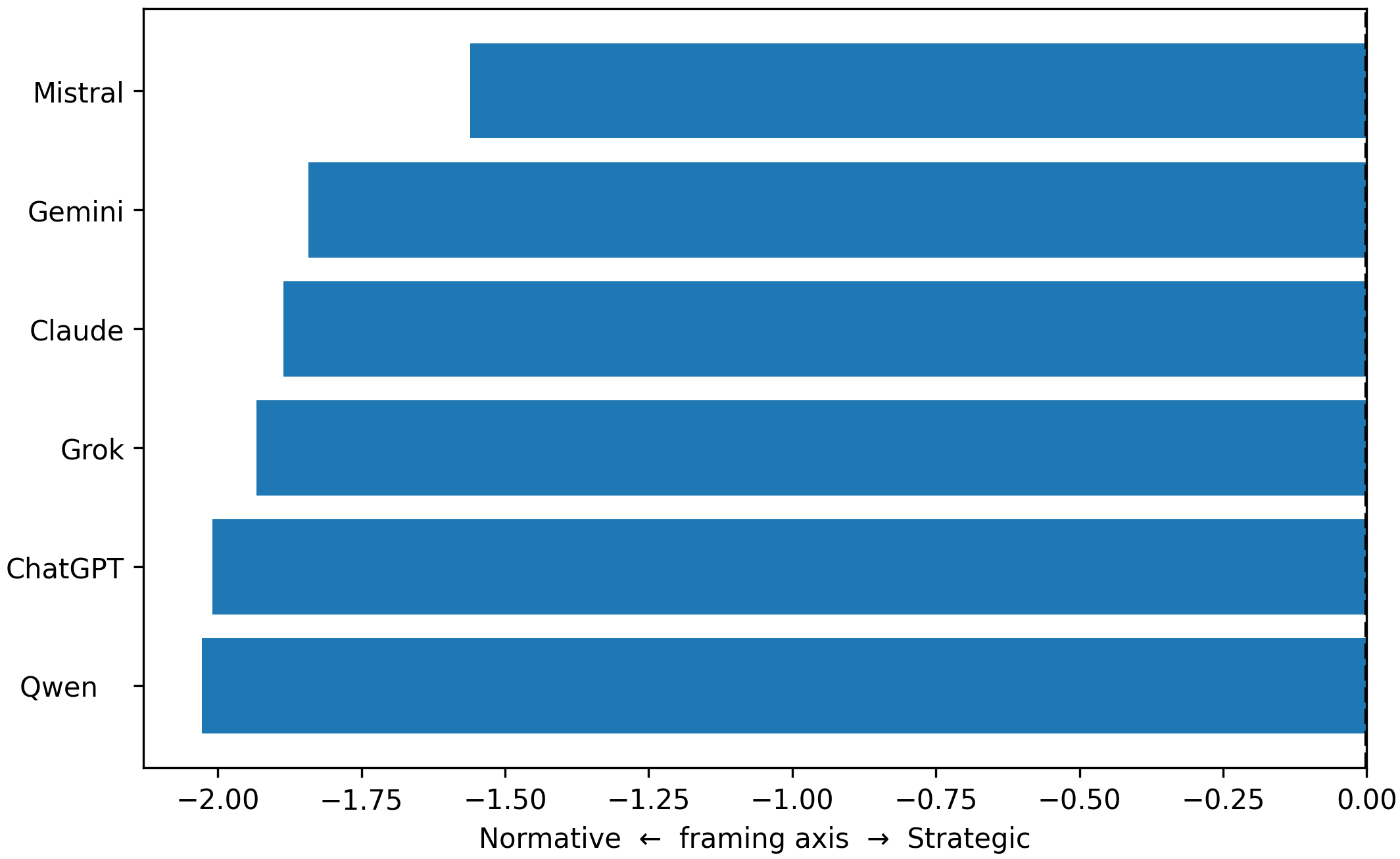}
    \caption{One-dimensional framing orientation (normative/cooperative $\rightarrow$ strategic/deterrence).}
    \label{fig:framing-1d}
\end{subfigure}

\vspace{0.6em}

\begin{subfigure}[t]{0.48\linewidth}
    \centering
    \includegraphics[width=\linewidth]{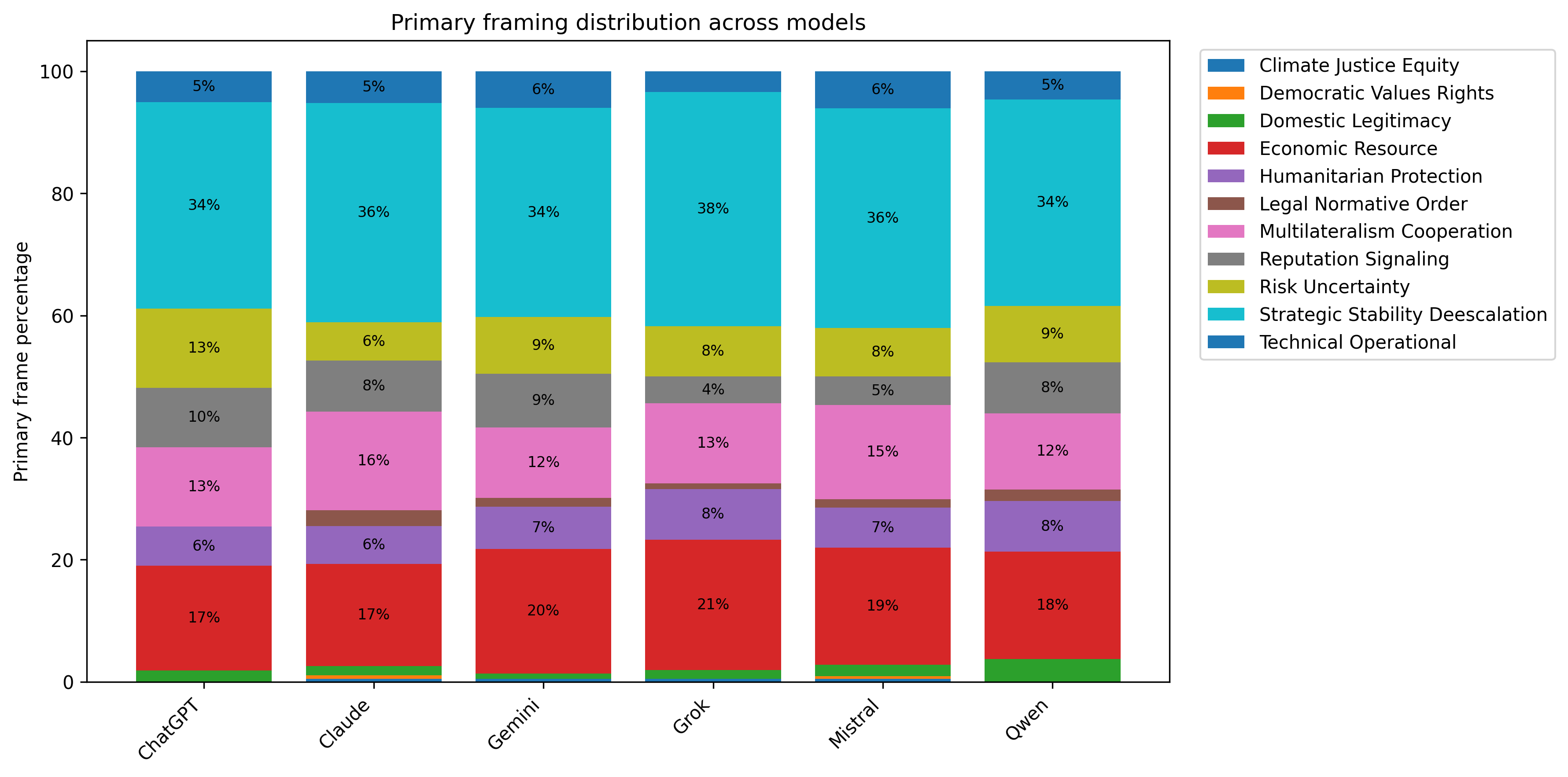}
    \caption{Primary frame distribution across models.}
    \label{fig:app-primary-framing}
\end{subfigure}
\hfill
\begin{subfigure}[t]{0.48\linewidth}
    \centering
    \includegraphics[width=\linewidth]{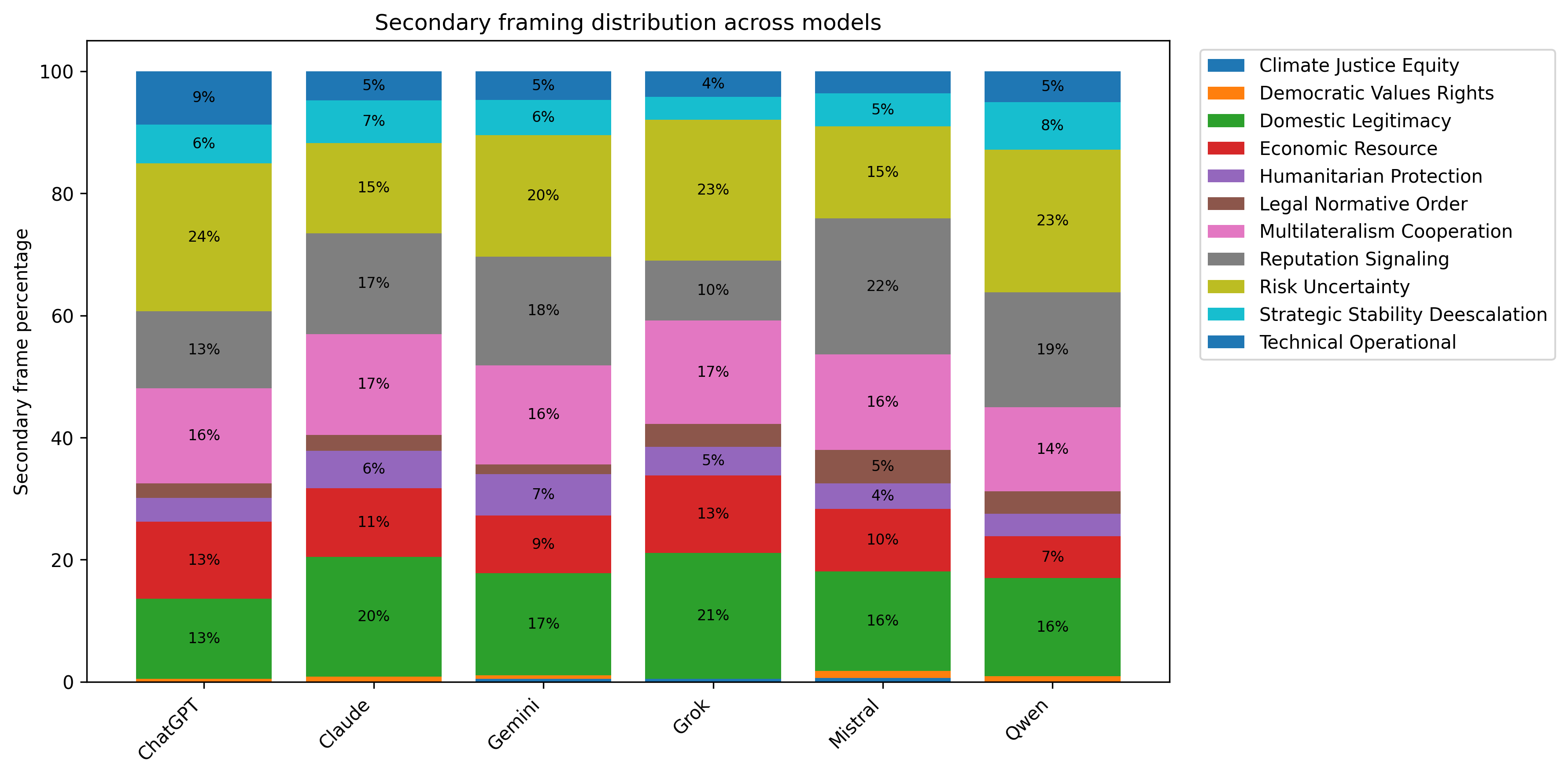}
    \caption{Secondary frame distribution across models.}
    \label{fig:app-secondary-framing}
\end{subfigure}

\caption{Supplementary framing and alignment results. Top: alignment by scenario and one-dimensional framing orientation. Bottom: primary and secondary frame distributions across models.}
\label{fig:app-framing-alignment-panel}
\end{figure*}

\clearpage
\subsection{Statistical tests}
\begin{table*}[h]
\centering
\caption{Mann-Whitney U tests comparing severity distributions between Round 1 and Round 2 for each model. Effect size reported as rank-biserial $r$. FDR correction applied across models.}
\label{tab:mw-round-shift}

{\footnotesize
\begin{tabular}{lrrrrc}
\toprule
Model  & $U$ & $p_{\text{raw}}$ & $r$ & $p_{\text{FDR}}$ & Sig. \\
\midrule
Mistral & 345.5 & $4.30\times10^{-12}$ & 0.756 & $2.58\times10^{-11}$ & Yes \\
Grok     & 533.0 & $4.13\times10^{-7}$  & 0.577 & $1.24\times10^{-6}$  & Yes \\
Gemini   & 671.5 & $1.29\times10^{-6}$  & 0.526 & $2.59\times10^{-6}$  & Yes \\
ChatGPT  & 676.5 & $2.30\times10^{-6}$  & 0.515 & $3.45\times10^{-6}$  & Yes \\
Claude   & 600.0 & $3.40\times10^{-6}$  & 0.524 & $4.08\times10^{-6}$  & Yes \\
Qwen     & 780.5 & $3.86\times10^{-5}$  & 0.449 & $3.86\times10^{-5}$  & Yes \\
\bottomrule
\end{tabular}
}
\end{table*}
\begin{table}[h]
\centering
\caption{Kruskal-Wallis tests assessing severity differences across decision dimensions (Economic, Security, Political) within each model. FDR correction applied across models. $n_{r1}=63$; $n_{r2}=45$ }
\label{tab:kw-dimension}

{\footnotesize
\begin{tabular}{lrrrrc}
\toprule
Model & $H$ & $p_{\text{raw}}$ & $p_{\text{FDR}}$ & Sig. \\
\midrule
Qwen     & 4.77 & 0.092 & 0.323 & No \\
Claude   & 0.80 & 0.669 & 0.884 & No \\
Grok     & 8.50 & 0.014 & 0.100 & No \\
Mistral  & 0.06 & 0.973 & 0.973 & No \\
ChatGPT  & 2.06 & 0.358 & 0.750 & No \\
Gemini   & 1.69 & 0.429 & 0.750 & No \\
Human    & 0.56 & 0.758 & 0.884 & No \\
\bottomrule
\end{tabular}
}
\end{table}

\begin{table}[h]
\centering
\caption{Post-hoc Mann-Whitney comparisons between Security and Political decisions. Only Grok shows a significant difference after FDR correction.}
\label{tab:posthoc}

{\footnotesize
\begin{tabular}{l l r r r r c}
\toprule
Model & Comparison & $U$ & $p_{\text{raw}}$ & $p_{\text{FDR}}$ & $r$ & Sig. \\
\midrule
Grok & Security vs Political & 387.0 & 0.0023 & 0.0493 & 0.40 & Yes \\
All others & --- & --- & --- & --- & --- & No \\
\bottomrule
\end{tabular}
}
\end{table}
\clearpage
\subsection{Framing annotation prompt}
\label{frame_explanations}
\begin{tcolorbox}[
  title={Framing explanations},
  colback=gray!5,
  colframe=gray!40,
  boxrule=0.6pt,
  arc=2pt,
  left=6pt,right=6pt,top=6pt,bottom=6pt
]
\footnotesize\ttfamily
\begin{verbatim}
You are annotating argumentation frames in decision explanations from geopolitical 
simulations.

FRAME DEFINITIONS (choose frames that describe the REASONING in the explanation):

1) Humanitarian_Protection
   - Focus: civilians/people, safety, harm reduction, relief, saving lives, protecting 
   vulnerable communities, humanitarian assistance, emergency response.

2) Risk_Uncertainty
   - Focus: uncertainty, incomplete info, contingency planning, mitigation, prevention, 
   resilience,
     risk assessment, avoiding worst-case outcomes, "we don't know yet".
3) Strategic_Stability_Deescalation
   - Focus: avoiding escalation, deterrence, signaling to adversaries, stability, balance 
   of power,
     preventing conflict spillover, security dilemma, strategic restraint.

4) Economic_Resource
   - Focus: costs/benefits, markets, supply chains, investment, economic growth/decline,
     resource allocation, fiscal constraints, economic incentives.

5) Legal_Normative_Order
   - Focus: international law, sovereignty, territorial integrity, UN processes, treaties,
     rules-based order, compliance/legitimacy from legal norms, sanctions rationale as 
     legal/norm-based.

6) Democratic_Values_Rights
   - Focus: democracy, freedom, civil liberties, human rights, rule of law as a VALUE,
     protection of democratic institutions, normative commitments 
     (“our values”, “liberal order”).

7) Multilateralism_Cooperation
   - Focus: alliances, partners, coordination, collective action, coalition-building,
     working through institutions (EU/NATO/UN/NGOs), burden-sharing.

8) Domestic_Legitimacy
   - Focus: voters/public opinion, mandates, political feasibility, social cohesion, 
   internal stability, legitimacy at home, election constraints, domestic
   backlash/acceptance.

9) Technical_Operational
   - Focus: practical implementation capacity, logistics, infrastructure, tools/technology,
   data/AI systems,  feasibility, operational readiness, timelines,
   execution details.

10) Reputation_Signaling
   - Focus: credibility, reputation, optics, “sending a message”, prestige, soft power,
     trust by partners, deterrence by credibility, narrative/image management.

11) Climate_Justice_Equity
   - Focus: fairness/equity, disproportionate burden on vulnerable groups, 
   responsibility/solidarity, just transition, adaptation fairness, 
   historical responsibility, equity across regions.
\end{verbatim}
\end{tcolorbox}

\begin{tcolorbox}[
  title={Framing annotation prompt},
  colback=gray!5,
  colframe=gray!40,
  boxrule=0.6pt,
  arc=2pt,
  left=6pt,right=6pt,top=6pt,bottom=6pt
]
\footnotesize\ttfamily
\begin{verbatim}
<Framing explanations>

OUTPUT FORMAT:
Return a JSON object with:
- primary_frames: array of 1-2 frame labels
- secondary_frames: array of 0-4 frame labels
- confidence: number between 0 and 1
- rationale_short: 1-2 sentences explaining why you chose the frames.

ITEM TO ANNOTATE:
Simulation: <SIMULATION>
Model: <MODEL>
Decision: <DECISION_TEXT>
Explanation:
<EXPLANATION_TEXT>
\end{verbatim}
\end{tcolorbox}
\subsection{Example data}

\begin{table}[h]
\centering
\caption{Round 1 action menu for the USA actor in the US--China--Taiwan simulation (titles and instructor-defined severity).}
\label{tab:uct-usa-r1-menu}
\footnotesize
\setlength{\tabcolsep}{4pt}
\begin{tabularx}{\linewidth}{l l X}
\toprule
\textbf{Category} & \textbf{Severity} & \textbf{Action option (Title)} \\
\midrule
Economic/Hybrid & Low & Impose sanctions on Chinese Communist Party members \\
Economic/Hybrid & Low/Medium & Pressure allies to limit trade with China \\
Economic/Hybrid & Medium & Reduce semiconductor tech exports to China by 10\% \\
Economic/Hybrid & Medium/High & Reduce semiconductor tech exports to China by 50\% \\
Economic/Hybrid & High & Fully halt semiconductor tech exports to China \\
\midrule
Security/Intel & Low & Maintain passive, peacetime activities \\
Security/Intel & Low/Medium & Conduct freedom-of-navigation operation in the Taiwan Strait \\
Security/Intel & Low/Medium & Conduct military exercises with South Korea/Japan/Australia in the East/South China Seas \\
Security/Intel & Medium & Conduct reconnaissance missions against China \\
Security/Intel & Medium/High & Conduct disruptive cyber operations against China \\
Security/Intel & High & Conduct military operations against China \\
\midrule
Diplomatic/Political & Low & Organize phone call with Chinese officials \\
Diplomatic/Political & Low/Medium & Send low-level delegation to China \\
Diplomatic/Political & Medium & Propose diplomatic summit \\
Diplomatic/Political & Medium/High & Announce an easing/lifting of trade restrictions/tariffs on Chinese goods \\
Diplomatic/Political & High & Announce a complete lifting of trade restrictions/tariffs on Chinese goods \\
\bottomrule
\end{tabularx}
\end{table}

\begin{table}[t]
\centering
\caption{Round 1 choices for the USA actor: human participant selection versus each LLM (titles only).}
\label{tab:uct-usa-r1-choices}
\scriptsize
\setlength{\tabcolsep}{3pt}
\begin{tabularx}{\linewidth}{l X X X X X X X}
\toprule
\textbf{Category} &
\textbf{Human} &
\textbf{ChatGPT} &
\textbf{Claude} &
\textbf{Gemini} &
\textbf{Grok} &
\textbf{Mistral} &
\textbf{Qwen} \\
\midrule
Economic/Hybrid &
Impose sanctions on Chinese Communist Party members &
Impose sanctions on Chinese Communist Party members &
Pressure allies to limit trade with China &
Reduce semiconductor tech exports to China by 10\% &
Reduce semiconductor tech exports to China by 50\% &
Pressure allies to limit trade with China &
Pressure allies to limit trade with China \\
\midrule
Security/Intel &
Conduct freedom-of-navigation operation in the Taiwan Strait &
Conduct freedom-of-navigation operation in the Taiwan Strait &
Conduct freedom-of-navigation operation in the Taiwan Strait &
Conduct freedom-of-navigation operation in the Taiwan Strait &
Conduct freedom-of-navigation operation in the Taiwan Strait &
Conduct military exercises with South Korea/Japan/Australia in the East/South China Seas &
Conduct military exercises with South Korea/Japan/Australia in the East/South China Seas \\
\midrule
Diplomatic/Political &
Propose diplomatic summit &
Organize phone call with Chinese officials &
Organize phone call with Chinese officials &
Organize phone call with Chinese officials &
Send low-level delegation to China &
Propose diplomatic summit &
Propose diplomatic summit \\
\bottomrule
\end{tabularx}
\end{table}
\end{document}